\documentclass[letterpaper,twocolumn,10pt]{article}
\usepackage{usenix2019_v3}

\usepackage{tikz}
\usepackage{amsmath}

\usepackage{filecontents}

\newcommand{\todo}[1]{\textcolor{red}{#1}}
\newcommand{\zaifeng}[1]{\textcolor{red}{Zaifeng: #1}}

\newcommand{\tool}[0]{JigsawRL}

\usepackage{algorithm}
\usepackage{algpseudocode}
\usepackage{makecell}
\usepackage{amsmath}
\usepackage{booktabs}

\begin{document}

\date{}

\title{\Large \bf \tool{}: 
Assembling RL Pipelines for Efficient LLM Post-Training
}

\author{
{\rm Zhengding Hu, Hehua Ouyang, Chang Chen, Zaifeng Pan, Yue Guan,}\\
{\rm Zhongkai Yu, Zhen Wang, Steven Swanson, Yufei Ding}\\
\textit{University of California, San Diego}
}


\maketitle

\begin{abstract}

We present \tool{}, a cost-efficient framework that explores \emph{Pipeline Multiplexing} as a new dimension of RL parallelism. \tool{} decomposes each pipeline into a \emph{Sub-Stage Graph} that exposes the intra-stage and inter-worker imbalance hidden by stage-level systems. On this abstraction, \tool{} resolves multiplexing interference through dynamic resource allocation, eliminates fragmented utilization by migrating long-tail rollouts across workers, and formulates their coordination as a graph scheduling problem solved with a look-ahead heuristic. On 4-64 H100/A100 GPUs across different agentic RL pipelines and models, \tool{} achieves up to 1.85$\times$ throughput over Verl on synchronous RL, 1.54$\times$ over StreamRL and AReaL on asynchronous RL, and supports heterogeneous pipelines with moderate latency trade-off.


\end{abstract}

\section{Introduction}

Reinforcement learning (RL)~\cite{ouyang2022rlhf} has become a standard technique for LLM post-training. Unlike the one-time process of creating a foundation model, RL is often applied iteratively to adapt a single pretrained model into various specialized versions. By continuously collecting feedback on specific task sets, RL allows researchers to branch a base model into different directions, such as enhancing complex reasoning~\cite{guo2025deepseekr1, shao2024grpo, jaech2024openai-o1}, agentic behaviors~\cite{li2025flow, chen2025marl, wang2025ragen}, or tool-usage capabilities~\cite{feng2025retool}. For example, within the Qwen model family (0.5–72B parameters) alone, the community has already published approximately 180,000 RL fine-tuned variants~\cite{qwen_ecosystem}.

\begin{figure}[t]
    \centering
    \includegraphics[width=\linewidth]{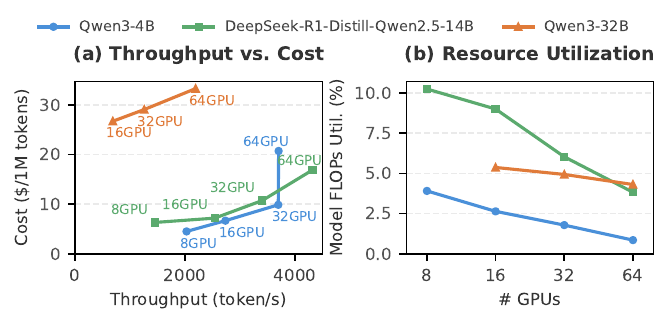}
    \caption{(a) Trade-off between pipeline throughput and monetary cost\protect\footnotemark across different models, spanning from base models~\cite{yang2025qwen3} to partially tuned variants~\cite{guo2025deepseekr1}. (b) Pipeline resource utilization, measured by FLOPs utilization (MFU).}
    \label{fig:intro_mfu}
\end{figure}

With such wide application, the \textbf{cost efficiency} of RL pipelines has emerged as a critical concern for both research and production. Yet, existing RL frameworks~\cite{sheng2025hybridflow, nemo-rl, slime_github} primarily optimize for time efficiency, while overlooking cost efficiency as a first-class objective. 
Our profiling on Verl~\cite{sheng2025hybridflow} reveals that improving time efficiency does not necessarily translate to better cost efficiency.
As shown in Figure~\ref{fig:intro_mfu}(a), scaling up GPU resources increases pipeline throughput, but leads to rapidly growing monetary costs with diminishing returns. This inefficiency stems from resource underutilization in RL pipelines. Figure~\ref{fig:intro_mfu}(b) demonstrates that the average GPU utilization, measured as MFU, is below 10\%. As the number of GPUs increases, MFU degrades even further.

\footnotetext{Monetary costs are estimated based on the on-demand pricing of AWS EC2 A100 instances~\cite{aws_ec2_pricing} (\$4.10 per GPU hour). }

\begin{figure*}[t]
    \centering
    \includegraphics[width=\linewidth]{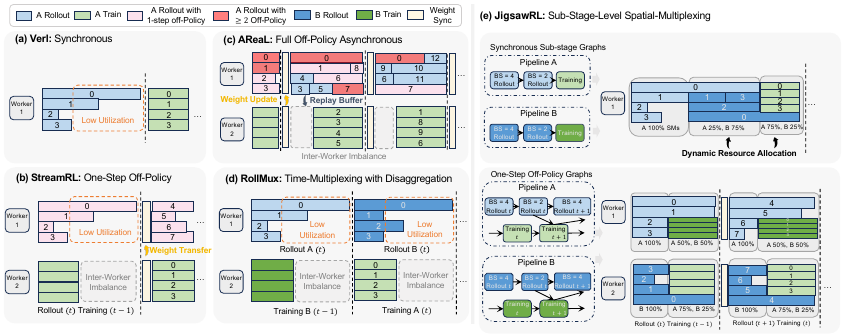}
    \caption{Comparison of execution behaviors under existing RL frameworks (a-d) and \tool{}'s sub-stage-level spatial multiplexing (e). Existing approaches suffer from low utilization and inter-worker imbalance. \tool{} multiplexes sub-stages with complementary resource demands from different pipelines to improve GPU utilization and thus cost efficiency.}

    \label{fig:intro_pipeline}
\end{figure*}




This cost inefficiency mainly comes from \textbf{workload imbalance} in the rollout stage~\cite{gao2025rollpacker, zhong2025stagefusion}. Due to the strict synchronization between alternating rollout and training, each step is gated by the slowest samples. As shown in Figure~\ref{fig:intro_pipeline}(a), a small number of long sequences dominate the rollout step, causing the effective batch size to shrink during decoding and thus reducing GPU utilization. This imbalance is further exacerbated by the emerging trend of agentic workloads, where multi-turn reasoning~\cite{chen2025marl, li2025flow} and external tool-calling~\cite{jin2025searchr1, feng2025retool} introduce highly stochastic and long-tailed sequence distributions.

Existing RL systems explore asynchronous execution~\cite{zhong2025streamrl, fu2025areal, sheng2025laminar} and time multiplexing~\cite{wu2025rollmux, wu2025rlboost, zhong2025stagefusion} to improve the efficiency, but they still fail to eliminate the workload imbalance both within and between stages. As shown in Figure~\ref{fig:intro_pipeline}(b), one-step off-policy execution such as StreamRL~\cite{zhong2025streamrl} overlaps rollout and training across workers, but the imbalance within rollout still remains. Figure~\ref{fig:intro_pipeline}(c) shows fully asynchronous execution like AReaL~\cite{fu2025areal}, which further relaxes synchronization but introduces imbalance across workers as rollout and training progress at different speeds. With these training and rollout disaggregation settings~\cite{zhong2025streamrl, sheng2025laminar, gao2025rollpacker}, partial resource reallocation helps mitigate imbalances across workers, but introduces additional overheads for frequent weight transfer. Moreover, off-policy methods introduce data staleness, which can affect convergence and stability~\cite{zheng2025staleness1}. Figure~\ref{fig:intro_pipeline}(d) shows time multiplexing such as RollMux~\cite{wu2025rollmux}, which enables concurrent pipelines with disaggregated workers, but imbalance within each rollout stage and across workers still exists.

The above works are still constrained by the dependencies and imbalance within a single RL pipeline, and GPU utilization still has room for improvement. In this work, we explore a new dimension of RL parallelism: \textit{Spatial Multiplexing of concurrent RL workloads}. We leverage multiple concurrent workloads in shared clusters and serverless environments~\cite{openpipe_serverless_rl, wei2026rlhfless, ye2026tensorhub}. This multi-workload setting provides more diverse stages and more concurrent execution opportunities, which better fills low-utilization periods. 

However, achieving efficient multiplexing is challenging due to the complex and dynamic execution behaviors in RL pipelines, making compute and memory resource demand vary over time. Naive approaches such as static resource partitioning~\cite{nvidia2022mps, nvidia2022mig} or reusing existing inference multiplexing frameworks~\cite{duan2024muxserve, yu2025prism} either lead to resource contention or fail to fill low-utilization periods effectively. Efficient multiplexing thus requires modeling these dynamic behaviors to enable resource coordination and pipeline scheduling.

To this end, we propose \tool{}, a cost-efficient and adaptive RL multiplexing system. The core abstraction is the \textit{Sub-Stage Execution Graph}, which decomposes coarse-grained stages (e.g., rollout and training) into fine-grained sub-stages with distinct compute and memory characteristics, exposing intra-stage and inter-worker imbalance that stage-level systems cannot see. Based on this abstraction, \tool{} formulates pipeline execution as two graph operations. \emph{Sub-stage Multiplexing} enables concurrent execution of sub-stages across pipelines with dynamic resource allocation. \emph{Sub-stage Merging} migrates long-tail rollout samples across DP instances to eliminate fragmented execution. A look-ahead heuristic coordinates both operations along the critical path.

In summary, our main contributions are as follows:
\begin{itemize}
    \item We propose the \textit{Sub-Stage Graph} abstraction that exposes intra-stage and inter-worker imbalance hidden by stage-level systems, turning pipeline multiplexing into a graph construction and scheduling problem.
    \item We observe that RL sub-stages exhibit complementary compute and memory demands, and exploit this through \emph{sub-stage multiplexing} with dynamic resource allocation.
    \item We observe that long-tail rollouts fragment GPU utilization across DP workers, and address this through \emph{sub-stage merging} with sample migration and balancing.
    \item On up to 64 GPUs across three agentic RL pipelines and six models, \tool{} achieves up to $1.85\times$ throughput over synchronous and $1.54\times$ over asynchronous baselines, and supports heterogeneous pipelines.
\end{itemize}

\section{Background and Related Work}

\subsection{Reinforcement Learning for LLMs}

Reinforcement learning (RL) has become an essential technique for aligning human preferences~\cite{ouyang2022rlhf, bai2022harmless} and enhancing capabilities~\cite{guo2025deepseekr1, jaech2024openai-o1}. Through iterative multi-stage training loops, RL enables models to obtain feedback from reward mechanisms and continuously refine their generation policies.
Therefore, optimizing the efficiency of RL pipelines is critical for meeting practical deployment demands.

Modern RL systems adopt a multi-stage pipeline design. A typical pipeline processes each global batch sequentially through several stages. The \emph{Rollout} stage generates responses from the current model. The \emph{Reference} stage computes log probabilities using a frozen model to regularize policy updates. The \emph{Reward} stage evaluates outputs using a reward model~\cite{ouyang2022rlhf} or rule-based signals~\cite{guo2025deepseekr1}. The \emph{Training} stage performs backpropagation and updates model parameters. Recent agentic pipelines~\cite{jin2025searchr1,feng2025retool} further include a \emph{Tool} stage to incorporate external feedback.

The performance bottleneck of RL pipelines varies across workloads. Table~\ref{tab:stage} shows the stage time breakdown of Verl~\cite{sheng2025hybridflow} under different models and tasks. We observe three key patterns. First, model capability and task difficulty affect stage composition: weaker models or harder tasks lead to longer reasoning chains and increase the proportion of rollout. Second, scaling GPUs shifts bottlenecks across stages. Due to limited scalability and sensitivity to imbalance, rollout becomes dominant at larger scales. Third, agentic workloads introduce additional overhead from tool usage, which can account for a large portion of execution time. Overall, these results show that bottlenecks are highly workload-dependent, requiring coordinated orchestration across stages.

\subsection{RL Frameworks}

Building efficient RL systems for LLMs has attracted considerable research attention. Existing frameworks~\cite{sheng2025hybridflow, nemo-rl, qin2025seer, mei2025real, yao2023deepspeedchat, hu2024openrlhf, wang2025roll, gao2025rollpacker, gao2025rollart, wu2025rollmux, slime_github, zhong2025stagefusion, he2026history, wu2025rlboost} adopt a tightly coupled design that leverages high-performance LLM serving engines~\cite{kwon2023vllm, zheng2024sglang} for the rollout stage and training engines~\cite{shoeybi2019megatron, zhao2023pytorch} for the training stages, while incorporating various optimizations for complex execution pipelines and stage orchestration.

\noindent \textbf{Parallelism Optimization for Synchronous RL.}
Standard RL pipelines operate in a strictly synchronous and sequential manner, alternating between rollout and training stages. Frameworks such as Verl~\cite{sheng2025hybridflow} and NeMo-RL~\cite{nemo-rl} allow independent parallelism configurations for the two engines by dynamically switching execution contexts at runtime and performing parameter resharding. ReaL~\cite{mei2025real} further formulates multi-stage pipeline deployment as a graph search problem, automatically identifying optimal parallelism configurations. Disaggregated frameworks~\cite{wang2025roll, hu2024openrlhf, tan2026orchestrrl} decouple rollout and training engines across separate GPU clusters to avoid step-wise context switching, but incur additional overheads such as GPU idling and communication for weight synchronization. Specific pipeline optimizations have also been proposed, including rollout length-based scheduling~\cite{gao2025rollpacker, qin2025seer} and speculative decoding~\cite{liu2025specRL, wang2025rlhfspec, he2026history, shao2025beat, chen2025respec}, to mitigate rollout imbalance.

\begin{table}[t]
\centering
\caption{Stage time proportion across different RL pipeline configurations with GRPO~\cite{shao2024grpo}.In our evaluated workload, the Reward stage is rule-based~\cite{guo2025deepseekr1} with negligible overhead.}
\label{tab:stage}
\scalebox{0.72}{
\setlength{\tabcolsep}{6pt}
\begin{tabular}{llrcccc}
\toprule
\makecell[c]{Model Size} & \makecell[c]{Task} & \makecell[c]{\#GPUs} & \makecell[c]{Rollout} & \makecell[c]{Reference} & \makecell[c]{Training} & \makecell[c]{Tool} \\
\midrule
Qwen3-0.6B         & GSM8K             & 4  & 66.1\% & 14.2\% & 19.7\% & ---    \\
Qwen3-4B           & GSM8K             & 4  & 52.2\% & 18.7\% & 29.1\% & ---    \\
Qwen3-4B           & AIME              & 4  & 80.9\% &  6.9\% & 12.2\% & ---    \\
Qwen3-4B           & GSM8K             & 64 & 67.4\% & 15.8\% & 16.8\% & ---    \\
Qwen3-4B           & Search-R1         & 4  & 39.1\% & 16.6\% & 24.6\% & 19.7\% \\
\bottomrule
\end{tabular}
}
\end{table}



\noindent \textbf{Asynchronous RL.} Synchronous RL suffers from imbalanced rollout (see Section~\ref{subsec:rollout_imbalance}) due to long-tail samples, which leads to underutilized GPU resources. Off-policy algorithms~\cite{noukhovitch2024asynchronousrl, ritter2026offpolicyagent} break the strict dependency between stages, enabling asynchronous rollout and training so that idle GPUs can proceed with subsequent-stage computation instead of blocking.
\textit{AReaL}\cite{fu2025areal} improves pipeline efficiency by discarding overlong samples and recomputing them later to mitigate the long-tail effect. \textit{Laminar}\cite{sheng2025laminar} proposes fully asynchronous rollout and trainer instances to break barriers between stages, leveraging relay buffers to support fine-grained weight updates and isolate long-tail samples.
\textit{RLinf}\cite{yu2025rlinf} enables more flexible data and stage partitioning at a finer granularity, achieving dynamic spatiotemporal scheduling within a single RL pipeline.
However, asynchronous RL suffers from data staleness\cite{zheng2025staleness1}, which can degrade training stability and convergence. Such trade-offs are undesirable in many real-world deployments, where strict correctness and stability requirements must be met, making it still critical to address inefficiencies within synchronous RL pipelines.

Overall, existing works focus on optimizing an \textit{individual RL training pipeline} for a single LLM.
In contrast, this paper explores a new dimension by \textit{multiplexing RL pipelines} to improve resource utilization. This introduces additional flexibility by enabling stages with diverse resource demands to share GPU resources, while remaining orthogonal to existing single-pipeline optimization methods.

\section{Motivation}

\subsection{RL Workload Imbalance}
\label{subsec:rollout_imbalance}

\begin{figure}[t]
    \centering
    \includegraphics[width=\linewidth]{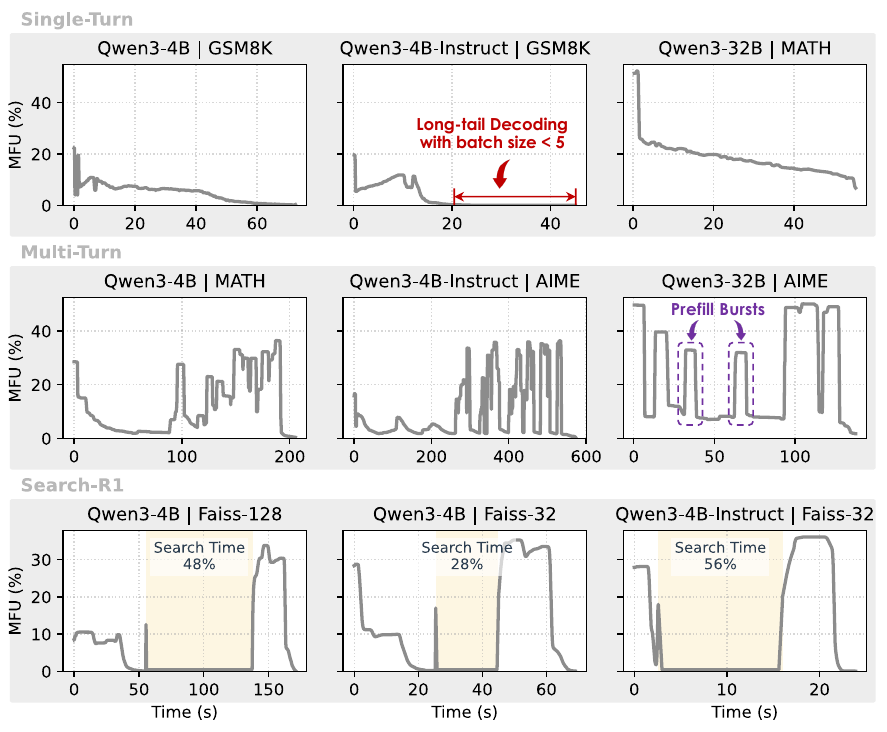}
    \caption{Different agentic behaviors in rollout stages and the resulting highly dynamic and imbalanced GPU utilization.}
    \label{fig:moti_imbalance}
\end{figure}

RL training exhibits highly imbalanced and dynamic GPU utilization patterns, primarily driven by the rollout stage. We profile the rollout GPU utilization overtime with GRPO~\cite{shao2024grpo} across different datasets and models, as shown in Figure~\ref{fig:moti_imbalance}. This imbalance comes from diverse agentic behaviors, including variation in single-turn decoding lengths, interleaved prefill and decoding in multi-turn interactions, and GPU idle with external tool usage. 

\noindent \textbf{Varied Decoding Lengths}. 
Decoding lengths are highly imbalanced across samples within a batch~\cite{gao2025rollpacker}, leading to a long-tail effect where a small number of long sequences determine the overall latency. As shorter sequences finish early, the effective batch size quickly shrinks, leaving the GPU underutilized for a large portion of the execution.

We also find that the imbalance is closely associated with model capability and task difficulty. Easier tasks or stronger models tend to produce shorter responses on average, but also more skewed length distributions. For example, under the same model size (Qwen3-4B), the more capable instruct-tuned variant (Qwen3-4B-Instruct) produces shorter responses on GSM8K, but with a more skewed length distribution, causing the GPU to operate at very small effective batch sizes for over 50\% of the decoding time. 

\noindent \textbf{Multi-turn Rollout Idling}. 
In agentic RL workloads, sampling exhibits multi-turn iterative behaviors~\cite{wang2025ragen, wang2025vagen, shani2024multiturnrlhf, abdulhai2023multi-turn-rl-bench, li2025flow}. This leads to an interleaving pattern of prefill and decoding phases: Prefill is no longer a one-shot initialization stage. Instead, after a period of decoding, new turns trigger additional prefill requests. Thus, GPU utilization exhibits a periodic rise-and-fall, with sharp utilization spikes during compute-intensive prefill bursts, followed by lower, memory-bound utilization during decoding.

We also observe imbalanced distribution in the number of turns across different samples: while most samples complete within 1--2 turns, a small fraction may extend to 4--5 turns. Such skewness further amplifies utilization imbalance.

\noindent \textbf{GPU Idle with Tool Usage}. In agentic RL pipelines~\cite{jin2025searchr1, feng2025retool, shang2025rstar2}, coordinating LLM execution with external tools introduces additional utilization imbalance. Tool usage typically runs on CPUs or relies on remote services, during which GPUs remain idle while waiting for responses. As tool latency increases (e.g., in RAG workloads), such idle periods can account for over 50\% of the total execution time. 

Tool usage also introduces new context, triggering additional prefill bursts. We observe that tool outputs are themselves highly imbalanced. For example, retrieved document lengths in RAG can vary by over 50\%, which further amplifies sequence length imbalance across requests.

Moreover, blocked tool calls introduce long-tail effects due to the extra synchronization, as execution must wait for all requests to generate their tool calling instructions. Blocked calling is common for heavy tools (e.g., database queries~\cite{douze2025faiss, hu2025hedrarag}, sandboxed execution~\cite{yao2022webshop, zhou2023webarena, jimenez2023swebench}, or search APIs~\cite{qin2023toolllm, patil2024gorilla}), where single-batched tool execution is more efficient.

\begin{figure}[t]
    \centering
    \includegraphics[width=\linewidth]{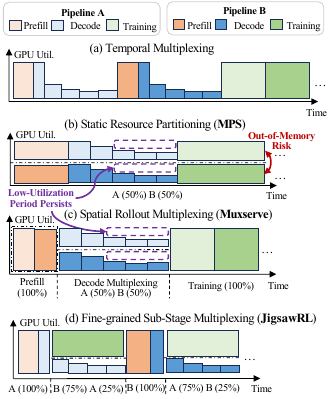}
    \caption{Comparison across different multiplexing methods.}
    \label{fig:multiplexing}
\end{figure}

\begin{figure*}[t]
    \centering
    \includegraphics[width=\linewidth]{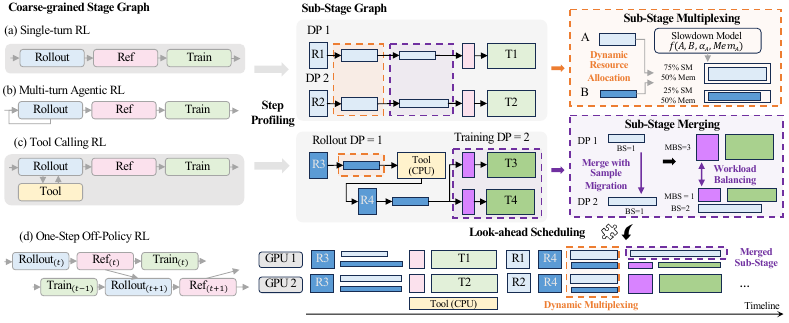}
    \caption{Overview of \tool{}. Starting from coarse-grained pipeline graphs, \tool{} constructs fine-grained sub-stage graphs, and enables efficient concurrent pipeline execution through sub-stage multiplexing, merging and graph scheduling.}
    \label{fig:overview_graph}
\end{figure*}

\subsection{Challenges in RL Multiplexing}

\label{sec:multiplex challenge}

Given the highly imbalanced and dynamic GPU utilization observed in RL pipelines, we explore an additional optimization dimension: \textit{Improve utilization by multiplexing different RL pipelines}. By multiplexing pipelines, stages with different utilization characteristics (e.g., compute-intensive prefill/training and memory-bound long-tail decoding) can share the GPU resources, thereby improving overall resource utilization and cost-efficiency. Such opportunities commonly arise in practical settings where RL workloads from multiple users are executed concurrently on shared GPU clusters, cloud servers, or serverless environments~\cite{openpipe_serverless_rl, thinkingmachines_tinker}.

However, achieving efficient multiplexing for RL pipelines remains challenging. As illustrated in Figure~\ref{fig:multiplexing}, existing approaches are insufficient to handle the complex and dynamic stage orchestration. Temporal multiplexing (a) executes different stages sequentially, and thus fails to utilize idle resources within each stage. Spatial multiplexing (b), e.g., via MPS~\cite{nvidia2022mps} or MIG~\cite{nvidia2022mig}, enables concurrent execution of multiple pipelines, but naively multiplexing full pipelines can lead to severe memory pressure. Memory-consuming training stages may overlap and cause out-of-memory (OOM) failures. Directly adopting LLM inference multiplexing systems~\cite{duan2024muxserve} enables spatial rollout multiplexing (c). However, concurrent execution of the long-tail, memory-bound rollout stage still leads to low GPU utilization. Meanwhile, the training stage remains serialized because of the huge memory consumption, resulting in significantly increased per-pipeline latency. 

In contrast, \tool{} (d) enables more fine-grained multiplexing by explicitly identifying and co-scheduling resource-complementary stages, while dynamically adapting resource allocation. This design effectively mitigates both resource contention and long-tail inefficiencies, achieving a better balance between overall throughput and pipeline latency.

\section{System Design}


We propose \tool{}, a cost-efficient and adaptive RL multiplexing framework. \tool{} introduces a graph-based abstraction. This abstraction models the RL pipeline into fine-grained sub-stages to capture utilization imbalance. Based on this abstraction, \tool{} formulates pipeline multiplexing as two graph operations to improve resource utilization and mitigate contention. \textit{Sub-stage Multiplex} enables concurrent execution of complementary sub-stages across pipelines through dynamic resource allocation. \textit{Sub-stage Merge} aggregates low-utilization long-tail samples across DP instances to eliminate fragmented execution. \tool{} finally applies a heuristic look-ahead scheduling algorithm to coordinate these operations and construct efficient pipeline execution.


\subsection{Sub-Stage Graph Construction}
Proper abstraction granularity is fundamental for efficient end-to-end scheduling. 
Existing RL pipeline modeling methods~\cite{mei2025real, sheng2025hybridflow} operate at the stage level, where rollout, training, and reference forward are abstracted as atomic stages. This granularity hides intra-stage behavior. A single rollout stage contains high-utilization prefill bursts, low-utilization decoding and idles for external tool usage. Stage-level scheduling cannot distinguish them. Resource imbalance thus becomes invisible, and scheduling opportunities are lost.

\begin{figure}[t]
    \centering
    \includegraphics[width=\linewidth]{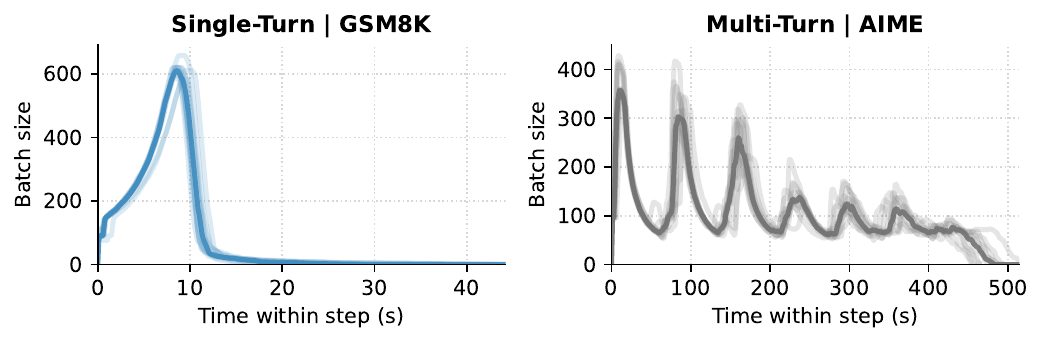}
    \caption{Decoding batch size variation across 10 adjacent steps in different pipelines using Qwen3-4B-Instruct, showing strong temporal consistency across steps.}
    \label{fig:bs_adjacentstep}
\end{figure}

\noindent \textbf{Sub-Stage Graph Abstraction}. \tool{} models the RL pipeline as a sub-stage graph. As shown in Figure~\ref{fig:overview_graph}, the graph starts from a stage-level structure (rollout, training, reference inference, tool calling), covering a range of agentic RL pipelines~\cite{guo2025deepseekr1, wang2025vagen, jin2025searchr1, fu2025areal}. \tool{} expands it into the \textit{sub-stage graph} to expose both intra-stage and inter-worker imbalance.
\begin{itemize}
    \item \noindent \textit{Intra-stage}: We decompose the rollout stage based on the number of tokens processed per forward computation, including prefill tokens and active decoding requests. Token count determines computational density, so sub-stages defined this way have uniform resource profiles.
    \item \noindent \textit{Inter-worker}: Different DP workers process heterogeneous samples. The sub-stage graph contains one replica per DP instance. Each replica tracks the sample subset assigned to its instance.
\end{itemize}



For each sub-stage, we need to profile for execution duration, memory footprint, batch size, and sequence length. These measurements guide multiplexing decisions. However, as discussed in \S\ref{subsec:rollout_imbalance}, different pipelines exhibit dynamic execution behavior, which makes offline construction difficult.

\noindent \textbf{Inter-Step Consistency}. Despite these dynamics, we observe strong temporal similarity in rollout behavior over adjacent steps. Figure~\ref{fig:bs_adjacentstep} shows the decoding batch size across 10 consecutive steps. In single-turn GSM8K, the degradation curves almost fully overlap. In multi-turn AIME, batch-size surges caused by repeated prefill bursts recur at similar positions. This temporal consistency arises because the training data distribution and the model's capability evolves slowly across adjacent steps. \tool{} therefore uses recent-step profiles to approximately construct and updates the sub-stage graph.

\noindent \textbf{Profile-based Graph Construction}. To instantiate the sub-stage graph, \tool{} extracts sub-stage boundaries from recent execution profiles. For each forward step, \tool{} computes its token count and maps it into workload buckets that correspond to distinct execution regimes with different compute and memory behavior. In practice, we use three buckets: $[0,128)$, $[128,1024)$, and $[1024,\infty)$. \tool{} updates the current sub-stage only when the bucket assignment remains stable over a short window. Let $B_k$ denote the bucket index of the $k$-th forward step. A transition is triggered only when the bucket remains unchanged for $L_s$ consecutive steps. In practice, we set $L_s = 10$, which is sufficient to suppress fluctuations while preserving persistent shifts between sub-stages.

\begin{figure}[t]
    \centering
    \includegraphics[width=\linewidth]{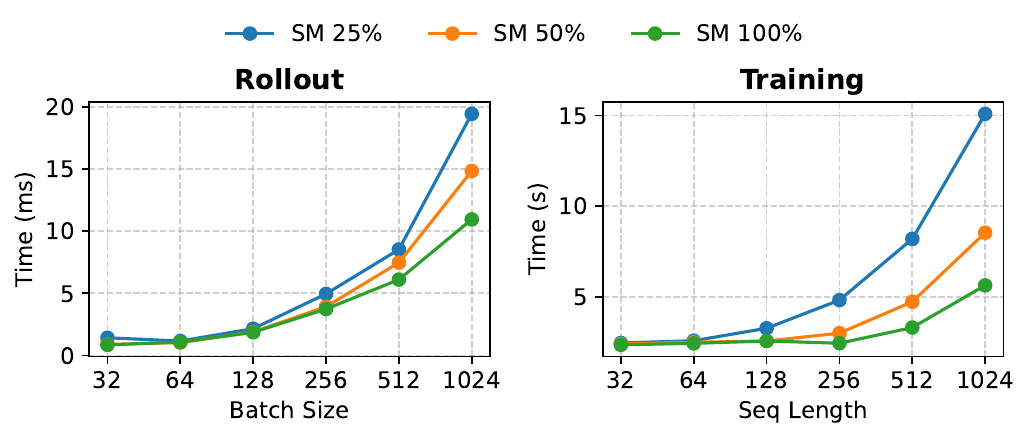}
    \caption{Latency comparison of Rollout and Training sub-stages when using different compute resources (SMs), evaluated on Qwen3-4B on 4 H100 GPUs.}
    \label{fig:kernel_charact}
\end{figure}

\subsection{Sub-stage Multiplexing}
\label{sec:subs-colocation}

\tool{} multiplexes sub-stages from different graphs to improve resource utilization. The concurrent execution of two sub-stages introduces two key constraints. First, the sub-stages compete for GPU compute resources, leading to potential slowdown due to SM contention. Second, their combined memory footprint must fit within the available GPU memory. \tool{} must adjust resource allocation at the sub-stage level to enable efficient multiplexing. Static and coarse-grained multiplexing frameworks~\cite{nvidia2022mps, nvidia2022mig, duan2024muxserve} fall short in this setting.

\begin{figure}[t]
    \centering
    \includegraphics[width=\linewidth]{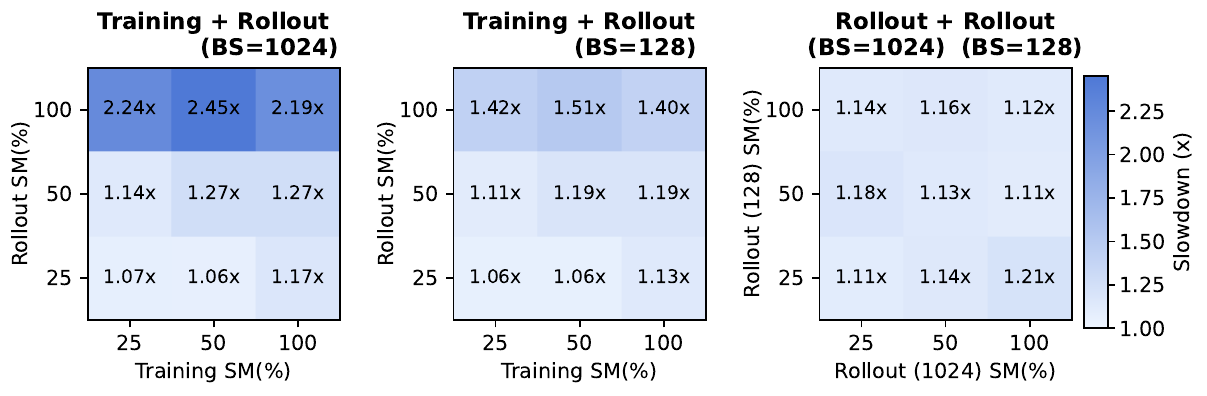}
    \caption{Slowdown under different SM partitioning when multiplexing two sub-stages (A + B). Each cell shows the slowdown of sub-stage A compared to isolated execution, evaluated on Qwen3-4B on 4 H100 GPUs.}
    \label{fig:heatmap}
\end{figure}

\noindent \textbf{Compute Resource Utilization}. Different sub-stages exhibit distinct sensitivity to GPU compute resources (SMs). We study this by controlling SM numbers used by the kernel via NVIDIA MPS~\cite{nvidia2022mps}. As shown in Figure~\ref{fig:kernel_charact}, compute-intensive sub-stages, including Training and large-batch Rollout ($\ge 256$), are highly sensitive to SM allocation, with slowdown up to 2.7$\times$ with 25\% SMs. In contrast, small-batch Rollout ($<128$) is largely insensitive to SM availability, indicating a memory-bound regime. These differences lead to asymmetric interference. As shown in Figure~\ref{fig:heatmap}, multiplexing compute-bound sub-stages results in strong contention (slowdown up to 2.19$\times$), while pairing compute-bound and memory-bound sub-stages yields better efficiency due to complementary resource usage. Based on this, \tool{} prioritizes multiplexing complementary sub-stages and applies SM partitioning to mitigate compute resource contention.

\begin{figure}[t]
    \centering
    \includegraphics[width=\linewidth]{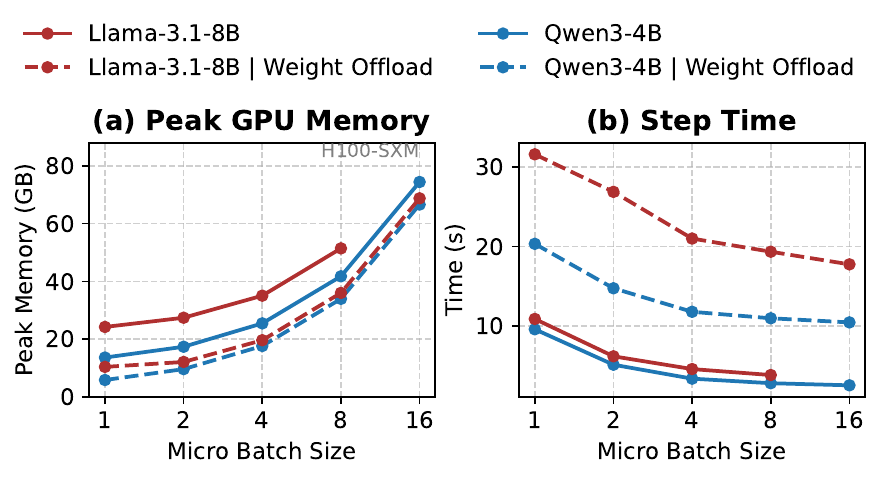}
    \caption{Impact of micro-batch size on per-GPU memory usage and training time on 4 H100 GPUs. Results are shown for both standard training and CPU weight offloading~\cite{rajbhandari2020zero}, with a fixed global batch size of 256.}
    \label{fig:mbs_memory}
\end{figure}

\noindent \textbf{GPU Memory Utilization}. Memory footprint directly constrains both execution efficiency and multiplexing feasibility. Training is more sensitive to memory availability. As shown in Figure~\ref{fig:mbs_memory}, during training, micro-batch size determines GPU memory usage and introduces a trade-off between throughput and memory consumption. During rollout, memory usage is largely driven by model weights and the KV cache, which bound the number of concurrent decoding requests. Under large-model settings (e.g., serving Qwen3-32B on 4 H100 GPUs), we observe up to 1.43$\times$ slowdown when reducing the memory consumption from 80\% to 40\%. Therefore, the GPU memory allocation for each sub-stage also becomes a key control knob for multiplexing.

\noindent \textbf{Slowdown Modeling}. Given the compute and memory constraints above, \tool{} needs to predict how much slowdown each multiplexing choice incurs. However, exhaustively profiling multiplexing slowdown over all execution parameters is impractical. \tool{} therefore constructs the slowdown model over a compressed space of sub-stage types and discrete resource budgets. We discretize the resource allocation space to keep profiling tractable. Specifically, compute allocation is discretized into four SM budgets (25\%, 50\%, 75\%, and 100\%), and memory allocation into four levels (20\%, 40\%, 60\%, and 80\% of the GPU memory budget). Combined with $\sim$5 sub-stage types, this results in on the order of $10^2$ measurements per pipeline, which is lightweight as an initialization cost. We denote the slowdown lookup function of sub-stage $k_A$ when co-located with $k_B$ as
\[
s(k_A, k_B) = f(k_A, k_B, \alpha_A, Mem_A),
\]
where $\alpha_A$ and $Mem_A$ denote the compute and memory share assigned to $k_A$, with the remaining resources assigned to $k_B$. In practice, configurations with similar $(\alpha_A, Mem_A)$ exhibit similar slowdown behavior, allowing interpolation between neighboring profiled points. \tool{} profiles $f$ offline on this sparse grid and interpolates to estimate unseen configurations.

\noindent \textbf{Dynamic Resource Allocation}. To reduce compute resource contention, we leverage NVIDIA Green Context~\cite{nvidia2025greencontext} to bound the SM resources available to individual sub-stages. Each process partitions the device SMs and creates streams that limit the maximum number of SMs according to the discrete budgets. 
For the rollout sub-stages, we pre-record multiple groups of CUDA graphs on these resource-bounded streams during inference server initialization~\cite{zheng2024sglang}; at each decoding step, the system replays the CUDA graph corresponding to the SM budget determined by the current co-location scenario. For the reference and training sub-stages~\cite{zhao2023pytorch}, we directly bind the resource-bounded stream to the forward and backward operations. This coarse-grained discretization keeps the CUDA graph memory footprint manageable and avoids the wave quantization effects that arise when high-performance kernels (e.g., GEMMs, FlashAttention) are mapped to irregular SM counts~\cite{duan2024muxserve, lu2025conco}.

For GPU memory resources, \tool{} dynamically adjusts the memory footprint to co-locate different sub-stages. For training and reference sub-stages, we tune the micro-batch size to trade off slowdown with memory usage. For rollout sub-stages, where memory allocation is fixed at launch time, \tool{} periodically reconfigures the KV cache pool based on recent execution behavior (every $\sim$30 steps in practice). The reconfiguration cost is amortized over multiple steps.

\subsection{Sub-stage Merging}
\label{sec:substagemerge}

\begin{figure}[t]
    \centering
    \includegraphics[width=\linewidth]{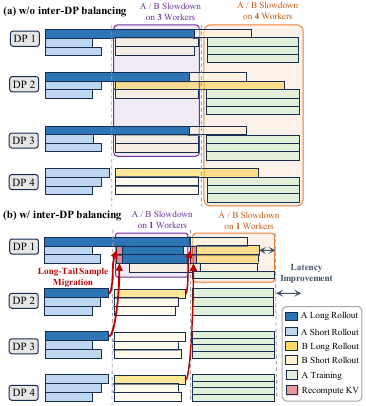}
    \caption{\tool{} mitigates inter-worker imbalance by sample migration and multiplexing-aware workload balancing.}
    \label{fig:inter_dp}
\end{figure}

Rollout across DP workers introduces additional workload imbalance that is not resolved by intra-DP sub-stage multiplexing. Long-tail samples may leave a subset of DP workers executing low-utilization rollout while others have already finished. As shown in Figure~\ref{fig:inter_dp}(a), even at low utilization, these sub-stages still consume compute and memory resources, slowing down other concurrent sub-stages.




\noindent \textbf{Long-Tail Sample Migration and Aggregation}. \tool{} handles such multiplexing inefficiency by dynamically migrating rollout samples across DP workers. As shown in Figure~\ref{fig:inter_dp}(b), \tool{} continuously tracks the execution progress of rollout sub-stages during model training. When multiple DP workers are detected to be executing low-utilization rollout sub-stages, \tool{} selects a target DP worker and migrates the long-tail rollout samples from other workers to it. This migration aggregates fragmented long-tail workloads onto a small number of DP workers. By leveraging the dynamic batching~\cite{yu2022orca} capability of the rollout engine, the aggregated workloads achieve higher GPU utilization with very small latency increase. Meanwhile, the rest of DP workers can release resources for the current pipeline, enabling more effective sub-stage execution for other pipelines.


\tool{} determines whether to perform migration based on the expected total execution time. Consider a set of low-utilization rollout sub-stages $\mathcal{K}=\{k_1, k_2, \dots, k_n\}$ on $n$ DP workers. Suppose DP instance $t$ is selected as the migration target. Each sub-stage $k_i$ is currently multiplexed with another sub-stage $k_i'$. We estimate the expected execution time improvement of migrating all rollout samples in $\mathcal{K}$ to the target workers $t$ as $\Delta T = T_{\text{origin}} - T_{\text{migr}}$, where

\begin{equation*}
T_{\text{origin}}
=
\max_{i\in[1,n]}
\left(
s(k_i,k_i')T(k_i)
\right)
+
\max_{i\in[1,n]}
\left(
s(k_i',k_i)T(k_i')
\right),
\end{equation*}

\begin{equation*}
T_{\text{migr}}=
T(\mathcal{K})
+
\sum_{\substack{i=1\\i\neq t}}^{n} M(k_i)
+
\max(
s(k_t',\mathcal{K})T(k_t'),
\max_{\substack{i\in[1,n], \\i\neq t}} T(k_i')
).
\end{equation*}


Here, $T(\mathcal{K})$ denotes the execution time of the aggregated rollout workload. $s(k_t', \mathcal{K})$ denotes the slowdown of $k_t'$ under multiplexing with the aggregated workload. $M(k_i)$ denotes the migration overhead of sub-stage $k_i$. In practice, the migration cost is dominated by KV cache recomputation. We estimate this overhead based on the FLOPs of the migrated samples and the average MFU of the prefill stages.



\noindent \textbf{Multiplexing-aware Workload Balancing}. After migrating and aggregating long-tail rollout samples of pipeline A onto a specific DP instance $t$, \tool{} further achieves workload balance across DP workers under multiplexing. When another pipeline B is under multiplexing, the uniform data parallelism or workload balancing in exclusive execution~\cite{zhao2023pytorch, wang2025wlb} becomes suboptimal. This is because DP worker $t$ may suffer from multiplexing slowdown, while other DP workers do not.

To address this, we introduce a multiplexing-aware load balancing strategy that dynamically skews the workload distribution. As illustrated in Figure~\ref{fig:inter_dp}, both the rollout sub-stages of B and training sub-stages of A assign less workload to DP worker $t$ to compensate for its multiplexing interference. Let $bs_i$ denote the batch size assigned to $DP_i$, and $T(bs)$ the exclusive execution time for a given batch size. To equalize execution time across all workers, we set:
$$ \frac{T(bs_i)}{T(bs_t)} \approx s(k_{B,t}, k_A), \quad \forall i \neq t, $$
where $s(k_{B,t}, k_A)$ is the slowdown factor estimated by the model in \S\ref{sec:subs-colocation}. In practice, when Pipeline B is in the rollout sub-stage, we achieve this by routing fewer requests to the inference server on $t$. For reference or training sub-stages, we dynamically adjust the micro-batch sizes across DP workers.

\begin{figure}
    \centering
    \includegraphics[width=\linewidth]{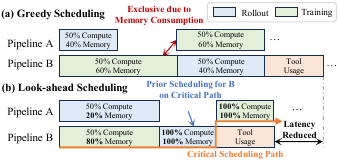}
    \caption{Comparison between greedy and critical-path-aware look-ahead scheduling.}
    \label{fig:scheduler}
\end{figure}

\begin{figure*}[t]
    \centering
    \includegraphics[width=\linewidth]{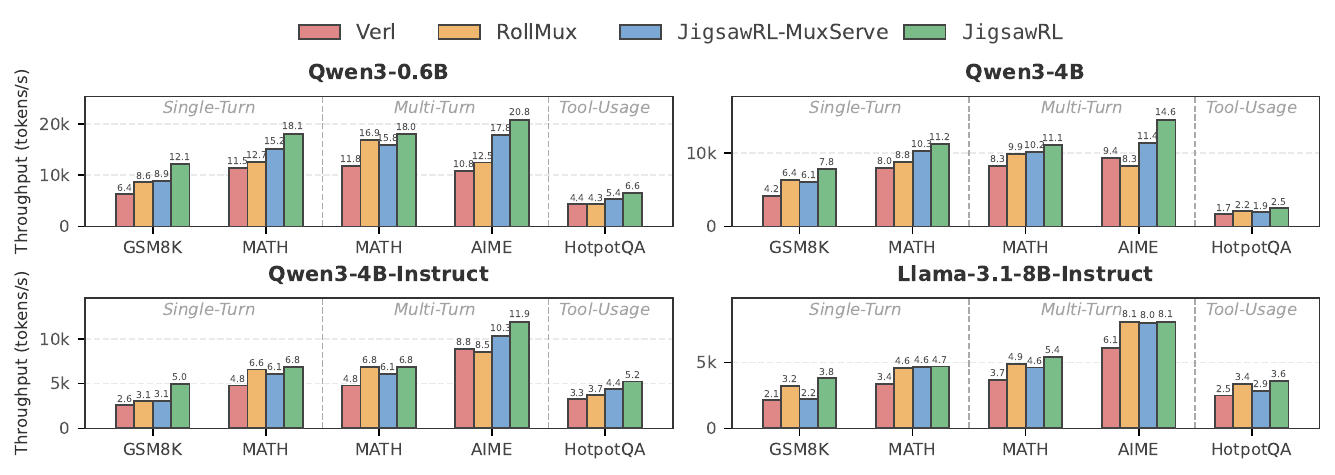}
    \caption{Throughput of homogeneous pipelines across different agentic RL pipelines on 8 H100 GPUs.}
    \label{fig:first_throughput}
\end{figure*}

\subsection{Look-Ahead Scheduling}
\label{sec:graph-scheduling}

With the sub-stage graph and slowdown model, \tool{} schedules sub-stages across co-located pipelines to minimize the total completion time. A greedy scheduler that minimizes only the immediate latency of the current co-location sub-stages can be globally suboptimal.
Figure~\ref{fig:scheduler} illustrates such an example. Greedy scheduling allocates more resources to Pipeline A’s rollout when multiplexing with Pipeline B’s training. However, this prevents A’s subsequent training from co-locating with B due to the memory constraints. In contrast, assigning fewer resources to A preserves this co-location opportunity. Moreover, since B’s rollout contains a tool-usage sub-stage, prioritizing its rollout allows the subsequent CPU-based tool calling to overlap with A’s training sub-stage, thereby further reducing the overall slowdown.

\tool{} adopts a windowed look-ahead policy when choosing among candidate scheduling actions. At each scheduling point, \tool{} maintains a frontier $\mathcal{F}$: the set of sub-stages that are ready to run across all co-located pipelines. On $\mathcal{F}$, \tool{} considers two operations in \S\ref{sec:subs-colocation} and \S\ref{sec:substagemerge}: (i) multiplexing a pair $(k_A, k_B) \in \mathcal{F}$ on the same worker under a resource split $r$, and (ii) merging two sub-stages on different workers into one. Each candidate action $a$ is thus paired with an allocation $r$, and its per-sub-stage execution time is derived from the slowdown model $s(k_A, k_B)$. Applying $(a, r)$ retires the chosen sub-stages and admits their successors along the intra-pipeline dependency edges; \tool{} then enumerates $(a, r)$ over the updated frontier and repeats, much like fitting puzzle pieces together one at a time.



The cost of $(a, r)$ is the critical path of $\mathcal{F}$, the longest chain of sub-stages that must execute sequentially:
\[
C(a,r)=\max_{p\in \mathcal{P}(\mathcal{F})} \sum_{k\in p} \hat{T}_{a,r}(k),
\]
where $\mathcal{P}(\mathcal{F})$ enumerates all dependency paths. This chain bounds how fast the window can finish, so minimizing $C(a, r)$ jointly accounts for the action and its allocation. \tool{} picks the $(a, r)$ with the smallest cost and continues scheduling. We use $W = 3$ by default, which we find offers a good balance between scheduling quality and search overhead.

\section{Evaluation}
\subsection{Experimental Setup}

\begin{figure*}[t]
    \centering
    \includegraphics[width=\linewidth]{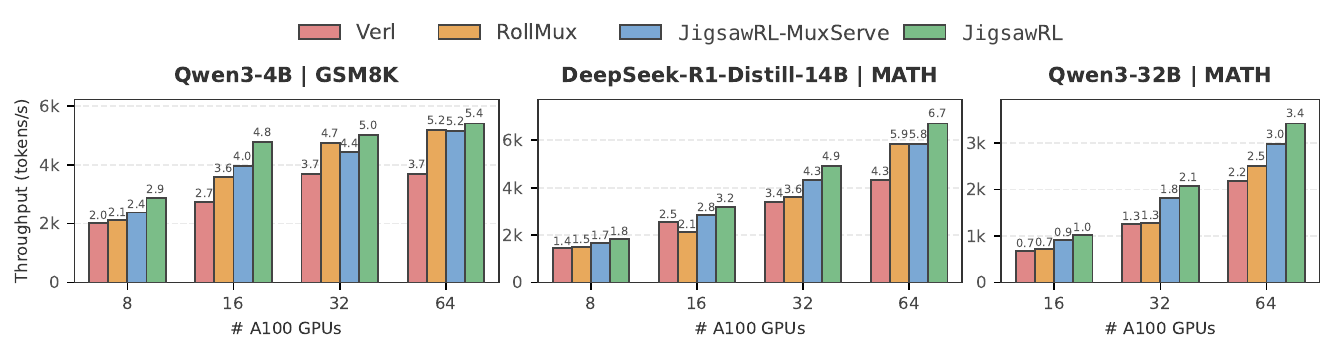}
    \caption{Throughput comparison between Verl and \tool{} scaling from 8 to 64 A100 GPUs with larger models.}
    \label{fig:scalability}
\end{figure*}

\noindent \textbf{Hardware Platforms}. We conduct experiments on both single-node and multi-node settings. For single-node evaluation, we use a cluster with 8 H100 GPUs, each with 80~GB HBM3 memory, interconnected via NVLink. For multi-node evaluation, we evaluate scalability on an A100 cluster with 64 GPUs, where each node contains 4 GPUs with 80~GB HBM memory. GPUs within each node are connected via NVLink.


\noindent \textbf{RL Pipelines}. 
We evaluate three representative RL pipelines for agent training. 
\textit{Single-Turn} generates a complete response in one step, commonly used for reasoning tasks~\cite{guo2025deepseekr1, jaech2024openai-o1}. 
\textit{Multi-Turn} iteratively produces intermediate outputs for self-refinement and multi-agent interaction~\cite{wang2025ragen, shani2024multiturnrlhf, chen2025marl, li2025flow}. 
\textit{Tool-Usage} invokes external tools during rollout; we use the Search-R1 pipeline~\cite{jin2025searchr1}. 
We adopt GRPO~\cite{shao2024grpo} for policy optimization, with a global batch size of 64, group size 4, and maximum response length of 8192 tokens.


\noindent \textbf{Datasets}. For single-turn pipelines, we use math and reasoning datasets, including GSM8K~\cite{cobbe2021gsm8k} and MATH~\cite{hendrycks2021math}. For multi-turn pipelines, we additionally include the challenging AIME dataset. For tool-usage pipelines, we use the multi-hop QA dataset HotpotQA~\cite{yang2018hotpotqa} for RAG-based agent training. We use the MS MARCO~\cite{bajaj2016ms} dataset as the external RAG database and build an IVF4096 ($nprobe$=32) index with FAISS~\cite{douze2025faiss}.


\noindent \textbf{Models}. 
We use base models including Qwen3-0.6B and Qwen3-4B, and instruct-tuned models including Qwen3-4B-Instruct and Llama-3.1-8B-Instruct. 
For larger-scale evaluation, we use DeepSeek-R1-Distill-14B, a Qwen2.5-14B model distilled from DeepSeek-R1~\cite{guo2025deepseekr1}, and the base model Qwen3-32B. 
These models span small- to medium-scale sizes with varying levels of fine-tuning and reasoning capabilities; in practice, instruct-tuned and distilled models exhibit stronger capabilities on the tasks in our test datasets.

\noindent \textbf{Baselines}. 
For synchronous RL, we use Verl~\cite{sheng2025hybridflow} and RollMux~\cite{wu2025rollmux} as baselines. Verl adopts a rollout--training with co-located workers on a same set of GPUs, while RollMux uses time-multiplexing with disaggregated workers. We also include a version of \tool{} using MuxServe\cite{duan2024muxserve} to multiplex the rollout stages, noted as \tool{}-MuxServe. 
For asynchronous RL, we use StreamRL~\cite{zhong2025streamrl} and AReaL~\cite{fu2025areal} as the baseline. StreamRL overlaps rollout and training stages on disaggregated workers with one-step bounded staleness. AReaL further adopts fully asynchronous execution, tolerating staleness up to several model update steps.

For backend implementations, we use SGLang~\cite{zheng2024sglang} for rollout and FSDP~\cite{zhao2023pytorch} for training, which are well-suited for agentic workloads and small- to mid-scale models. \tool{} also supports vLLM~\cite{kwon2023vllm} and Megatron~\cite{shoeybi2019megatron}. Existing frameworks such as Slime~\cite{slime_github} and NeMo~\cite{nemo-rl} follow similar designs by integrating state-of-the-art rollout and training engines.




\subsection{Synchronous Pipeline Settings}
We first evaluate the main deployment setting, where two synchronous RL pipelines with the same model architecture and datasets but different weights, are executed concurrently on shared GPUs. We define throughput as the aggregated number of tokens processed by all the pipelines per unit time.

\noindent \textbf{Overall Performance Improvement}.
We measure the throughput improvement of \tool{} over Verl across different agentic pipelines. As shown in Figure~\ref{fig:first_throughput}, \tool{} consistently achieves the highest throughput, with an average speedup of $1.56\times$ over Verl and up to $1.95\times$. Compared with RollMux and PuzzRL-MuxServe, \tool{} further improves throughput by $1.27\times$ and $1.23\times$ on average, respectively.

First, \tool{} achieves larger speedups over Verl with more severe rollout imbalance. For example, for the weaker model Qwen3-0.6B, the speedup on GSM8K is higher than on MATH, since the more difficult dataset tends to induce consistently long reasoning and generation, thus reducing imbalance and limiting the optimization opportunity.

Second, the speedups over RollMux are more strongly correlated with workload imbalance across stages. For example, in the multiturn AIME workload with Qwen3-4B, the rollout stage accounts for 71.1\%, leading to imbalance across workers. In contrast, for Llama-3.1-8B-Instruct, rollout and training are more balanced, with 56.3\% and 43.7\% of the total execution time, respectively. In such cases, RollMux and \tool{} achieve comparable throughput.

\begin{figure}[t]
    \centering
    \includegraphics[width=\linewidth]{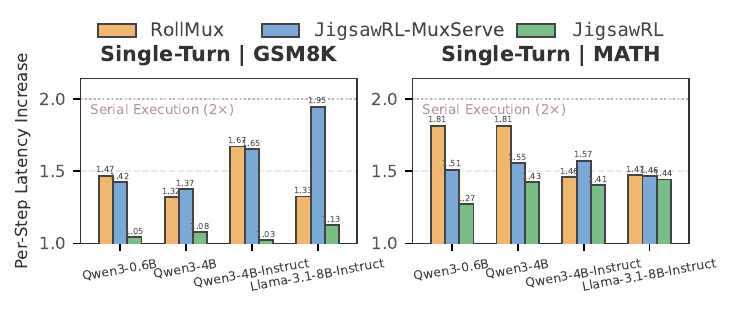}
    \caption{Latency increase of different multiplexing methods compared to the exclusive execution of Verl on 8 H100 GPUs. }
    \label{fig:latency_increase}
\end{figure}

\begin{figure}[t]
    \centering
    \includegraphics[width=\linewidth]{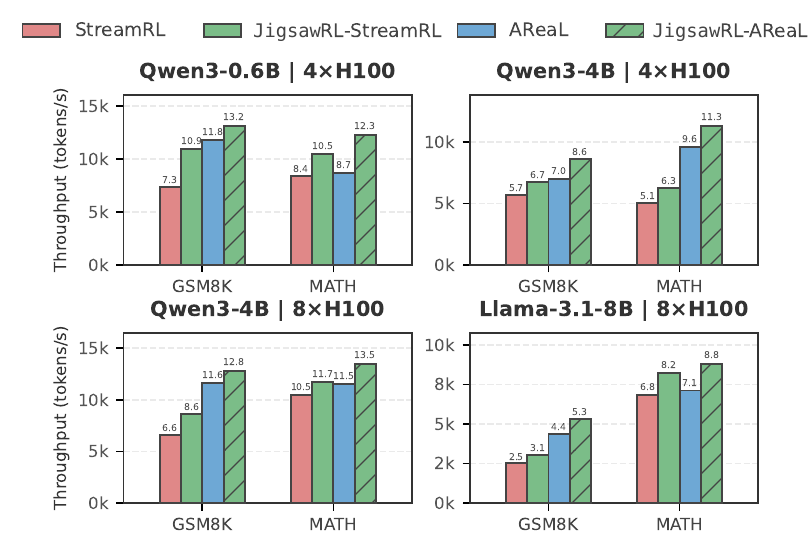}
    \caption{Throughput comparison of \tool{} under one-step off-policy (StreamRL) and fully asynchronous RL (AReaL).}
    \label{fig:async_throughput}
\end{figure}

\noindent \textbf{Scalability}. To evaluate \tool{} at scale, we conduct experiments on 8--64 A100 GPUs using larger models. As illustrated in Figure~\ref{fig:scalability}, \tool{} consistently scales across different model sizes and cluster sizes, achieving up to $1.75\times$ throughput improvement over Verl and RollMux with an average of $1.47\times$ and $1.31\times$, respectively. This scalability comes from two factors. First, as the cluster size increases, existing systems suffer from lower utilization and more pronounced stage imbalance, creating more opportunities for spatial multiplexing across complementary stages. Second, \tool{} further mitigates the long-tail effect at higher DP degrees through inter-DP-worker sample migration, which helps maintain balanced execution and sustain throughput gains at scale.


\noindent \textbf{Latency Trade-off}. While \tool{} improves aggregate throughput through multiplexing, it also increases the per-step latency of each individual pipeline due to multiplexed execution. Figure~\ref{fig:latency_increase} shows that this overhead remains moderate: the latency increase is $1.48\times$ on average and can be as low as $1.14\times$, substantially smaller than the $2\times$ latency cost of running two pipelines serially. We further observe that the latency increase is smaller on easier task sets such as GSM8K, which exhibit higher intra-batch imbalance.

\subsection{Diverse Pipeline Settings}
We further evaluate \tool{} on additional pipeline settings, including asynchronous pipelines and heterogeneous pipelines.

\noindent \textbf{Asynchronous RL}. Figure~\ref{fig:async_throughput} evaluates \tool{} under one-step off-policy~\cite{zhong2025streamrl} and fully asynchronous RL~\cite{fu2025areal}, using disaggregated rollout and training workers with equal resource allocation. \tool{} achieves an average speedup of $1.25\times$ (up to $1.54\times$) over StreamRL and $1.21\times$ (up to $1.41\times$) over AReaL. These results show that \tool{} remains effective in asynchronous settings, mitigating inter-worker imbalance and low utilization during decoding. Concretely, each worker enables multiplexing of the training stage of one pipeline with the rollout stage of another, thereby improving utilization.



\begin{figure}[t]
    \centering
    \includegraphics[width=\linewidth]{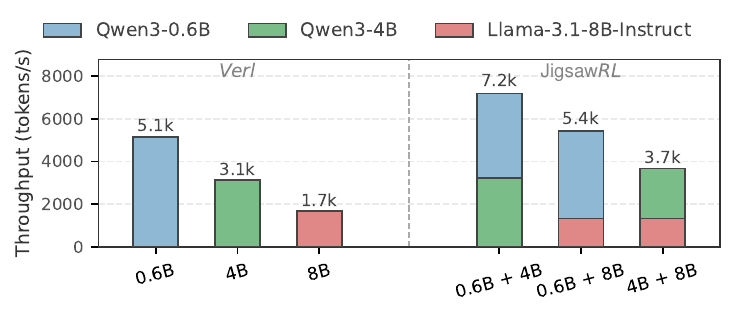}
    \caption{Throughput of heterogeneous pipelines in Verl and \tool{} on 4 H100 GPUs. The training dataset is GSM8K.}
    \label{fig:hetero_throughput}
\end{figure}

\noindent \textbf{Heterogeneous Models}. We also evaluate \tool{} with pipelines of different model sizes. As illustrated in Figure~\ref{fig:hetero_throughput}, \tool{} successfully improves the aggregated throughput with the individual pipeline slowdown limited to at most $1.34\times$. Interestingly, we observe a higher improvement when co-locating models have size disparities (Qwen3-0.6B and Llama-3.1-8B). This is because the sub-stages of the smaller model can better fit into the long-tail rollout sub-stages of the larger LLM and improve the utilization.

\begin{figure}[t]
    \centering
    \includegraphics[width=\linewidth]{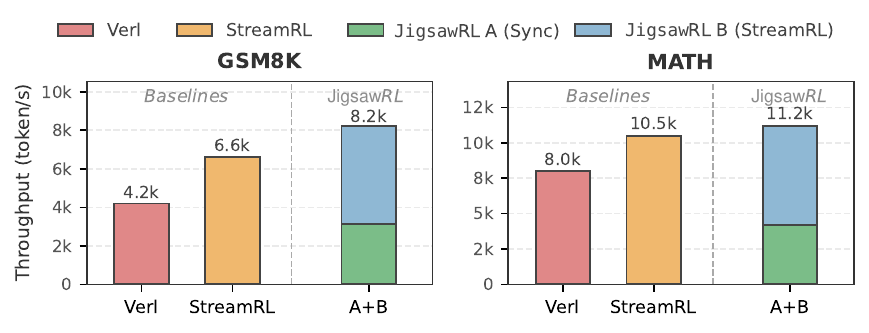}
    \caption{Throughput of \tool{} multiplexing synchronous (A) and one-step off-policy asynchronous (B) pipeline on 8 H100 GPUs. The model used is Qwen3-4B.}
    \label{fig:sync_async_throughput}
\end{figure}

\noindent \textbf{Synchronous and Asynchronous Pipelines}. We further evaluate \tool{} by multiplexing synchronous pipeline with an asynchronous one (StreamRL~\cite{zhong2025streamrl}). As illustrated in Figure~\ref{fig:sync_async_throughput}, \tool{} improves the aggregated throughput by $1.24\times$ over Verl and StreamRL on GSM8K, and $1.07\times$ on MATH. The improvement is higher on GSM8K because both pipelines exhibit more severe rollout imbalance under easier tasks, leaving more under-utilized periods that \tool{} can fill through sub-stage multiplexing. In contrast, MATH produces longer and more uniform rollouts in both pipelines, reducing the multiplexing opportunity.

\subsection{Case and Ablation Studies}

In this section, we study the compatibility of \tool{} with existing parallelism methods, as well as the effectiveness of individual optimization methods.

\begin{figure}[t]
    \centering
    \includegraphics[width=\linewidth]{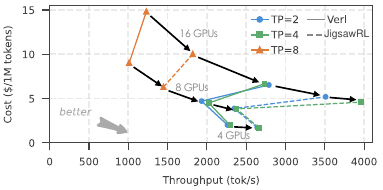}
    \caption{Throughput-Cost trade-off for Qwen3-4B scaling from 4 to 16 GPUs. The training dataset is GSM8K.}
    \label{fig:tp_comparison}
\end{figure}

\noindent \textbf{Integration with Other Parallelism Methods}. We evaluate \tool{} with Data Parallelism (DP) and Tensor Parallelism (TP). Figure~\ref{fig:tp_comparison} illustrates the throughput-cost trade-off space.

 In Verl, scaling with TP in a single node (TP=2 to TP=4) yields marginal gains, while cross-node scaling (TP=8) is limited by communication overhead due to the bandwidth gap between intra-node NVLink and inter-node networks. In contrast, DP is the more efficient parallelism for multi-node scaling: increasing from (TP=4, DP=1) to (TP=4, DP=4) achieves a $1.2\times$ throughput improvement, but incurs a $3.4\times$ cost increase due to exacerbated imbalance.

In contrast, our proposed Pipeline Multiplexing introduces a new, cost-efficient parallelism dimension. Under identical TP and DP settings, \tool{} improves the aggregated throughput by $1.16\times$ to $1.48\times$ without cost efficiency penalties. The flatter slope of its scaling curve demonstrates potential for larger-scale deployments. Furthermore, \tool{} is orthogonal to automated parallelism search frameworks like ReaL~\cite{mei2025real}, enabling further optimizations atop the ideal parallelism.

\begin{figure}[t]
    \centering
    \includegraphics[width=\linewidth]{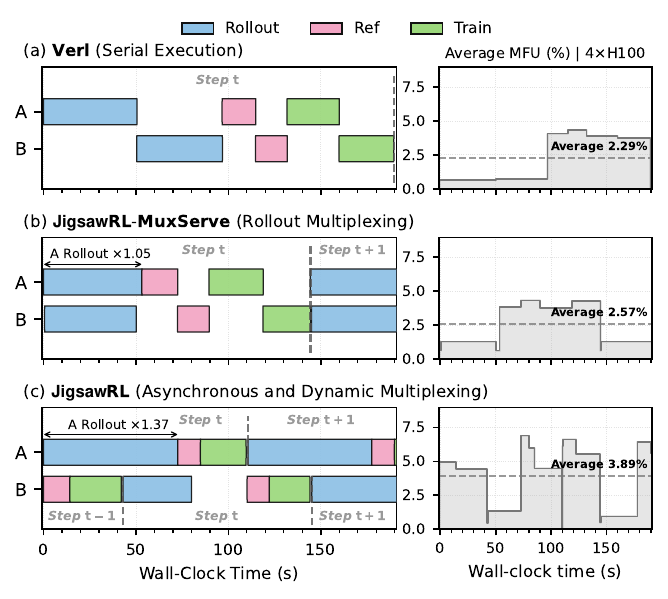}
    \caption{Pipeline multiplexing strategies and their impact on average MFU (Single-Turn, Qwen3-4B, 4×H100, GSM8K).}
    \label{fig:ablation_multiplex}
\end{figure}


\noindent \textbf{Dynamic Sub-Stage Multiplexing}. We evaluate the effectiveness of our sub-stage-level multiplexing by comparing different pipeline execution strategies, as shown in Figure~\ref{fig:ablation_multiplex}. Verl’s serial execution (a) exhibits prolonged low MFU during rollout due to long-tail effects. Using MuxServe for rollout multiplexing (b) partially improves overlap, but still suffers from low rollout utilization and increased latency due to serialized training. In contrast, \tool{} (c) enables asynchronous, fine-grained sub-stage multiplexing. By dynamically co-locating memory-bound rollout with compute-bound training and reference, it improves GPU utilization by $1.7\times$ and $1.5\times$ over Verl and \tool{}-MuxServe strategies, respectively.

\begin{figure}[t]
    \centering
    \includegraphics[width=\linewidth]{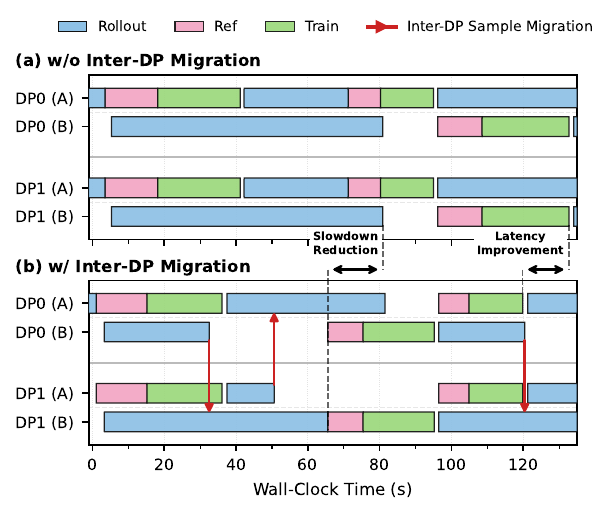}
    \caption{Inter-DP workload migration and its impact on execution latency (Single-Turn, Qwen3-4B, 4×H100, GSM8K)}
    \label{fig:inter_dp_eval}
\end{figure}

\noindent \textbf{Inter-worker Workload Migration and Balancing}. We evaluate the effectiveness of our long-tail sample migration and multiplexing-aware workload balancing mechanisms. As depicted in Figure~\ref{fig:inter_dp_eval}(a), without inter-DP migration, the long-tail samples of Pipeline A's rollout stage span across all DP instances. This execution continuously interferes with Rollout B. In contrast, in Figure~\ref{fig:inter_dp_eval}(b), by migrating the long-tail rollout samples of Pipeline A from DP1 to DP0, \tool{} effectively isolates the long-tail rollout sub-stage to different DP instances, alleviate the interference of sub-stage with extremely low utilization. Such optimization translates into the observable reduction in the overall training step latency improvement for both Pipeline A and B.

\section{Conclusion}

We present \tool{}, a cost-efficient RL post-training framework built on \emph{Pipeline Multiplexing}, a new dimension of RL parallelism. \tool{} models each pipeline as a sub-stage graph that exposes imbalance hidden by stage-level systems, and formulates pipeline multiplexing as a graph scheduling problem solved with a look-ahead heuristic. On up to 64 GPUs across three agentic RL pipelines and six models, \tool{} achieves up to $1.85\times$ throughput over synchronous baselines and $1.54\times$ over asynchronous baselines, while supporting heterogeneous pipelines with moderate latency overhead.

Our current study focuses on small- to mid-scale dense models, which remain the mainstream choice for agentic RL among individual researchers and smaller teams. Extending \tool{} to MoE architectures and LoRA-style parameter-efficient training opens further opportunities: MoE exposes expert-level imbalance that enables finer-grained multiplexing, while LoRA adapters over a shared base model naturally form multiplexing candidates with complementary memory footprints. We leave these directions to future work.



\bibliographystyle{plain}
\bibliography{ref}

\end{document}